# Technique and Challange for Multi-Camera Tracking

Wang Yong
College of Engineering & Information Technology
University of Chinese Academy of Sciences
Beijing China
wangyong@ucas.ac.cn

Lu Ke
College of Engineering & Information Technology
University of Chinese Academy of Sciences
Beijing China
luk@ucas.ac.cn

*Abstract*—**Multi-camera tracking is quite different from single camera tracking in mathematical principles and application scenarios, and it faces new technology and system architecture challenges. The existing theories and algorithms used in object matching, cameras calibration and topology estimation, and information fusion have been reviewed and show that the integrated application of multi techniques and multi theories is the key to solve the technology challenges. The distributed architectures of multi-camera tracking system based on camera processor and based on object agent have been compared and show that improving the computation ability of cameras and reducing the functions of control center is the key to solve the architecture challenges.**

*Keywords-multi-camera tracking; object matching; camera calibration; information fusion; system architecture*

## I. INTRODUCTION

Object tracking is an important topic in computer vision. The moving object often contains a large amount of visual information, as can provide a lot of valuable data for video analysis. Object tracking is often referring to object feature extraction, appropriate matching algorithm used to determine the object position, and other space-time change information to monitor objects in a video sequence. When tracking the moving objects, it often occur the phenomenon of object occlusion, temporarily disappearance, or into corners. Then it is easy to lose the object based on single camera. Multi-camera can observe the same scene from multiple angles, so it can provide more comprehensive information to solve this problem to some extent [1].

Due to the increase of cameras and visual angles, multi-camera tracking includes not only the knowledge of computer vision and information fusion, but also the theory of pattern recognition and artificial intelligence, and it has become a multidisciplinary research problem. Many researchers have explored the issues of multi-camera tracking and put forward some typical solutions. For example, Kang et al. [2] proposed to take the multi-camera tracking as a problem of maximum joint probability model based on color. By estimating the object model through Kalman filtering, it used the joint probabilistic data filtering and multi-camera homography to multi-object tracking. Nummiaro et al. [3] proposed a tracking algorithm for multi-view object based on particle filtering, but unlike the idea of information fusion, this algorithm selected the best point of view for object tracking among the different perspectives. Lien et al. [4] proposed a tracking method for multi-view object based on the cooperation of hidden Markov process and particle filtering.

The use of multiple cameras is helpful to solve the occlusion, chaotic scenes, the mutation of ambient light problems of moving object tracking, but due to the increase of cameras, the camera position relations and other concomitant factors, it also brings new technology and system architecture problems, such as object matching between multiple cameras, collaboration of cameras, automatic switching between different cameras, information fusion et al. The second part of this paper focused on the technical challenges of multi-camera tracking system and makes a review; the third part discussed the system architecture of multi-camera tracking and made a comparison of the distributed architectures based on camera processor and based on object agent; the fourth part made a summary of this paper, and the future developing directions were discussed.

## II. CHALLENGE OF TECHNOLOGY

Nowadays the key technical challenge of multi-camera tracking mainly concentrated in 3 interrelated parts, as are object matching, cameras calibration and topology estimation, and information fusion. (Shown in Fig. 1).

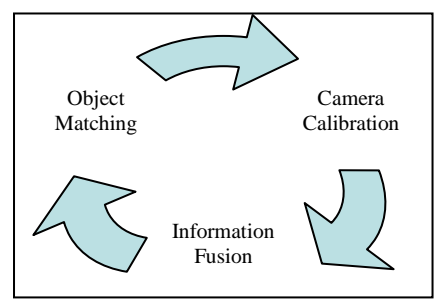

Fig. 1. Key technical challenge of multi-camera tracking

### A. Object Matching

Object Matching of multi-camera tracking mainly refers to matching the object in more than one camera view scope at the same time or different time, so as to locate and identify the object.

Set $x_i$ represent the feature vector of object $O$ in the $i$th camera, so with the time constraint $T$ the object matching

This work is supported by National Natural Science Foundation of China (#61371155).

function of object *O* in the whole *C* cameras can be written as following equation (1):

$$\arg\min_{x_i, x_j \in O} \sum_{t \in T} \sum_{i,j=1, i \neq j}^{C} \|x_i - x_j\|_t \quad (1)$$

In multi-camera tracking the object is different in color and shape in different cameras and the camera parameters and view angles are also different, so object matching in multi-camera tracking is greatly different from the traditional single-camera tracking. Object matching theory in single-camera tracking is no longer suitable for multi-camera tracking and it is necessary to establish a new theoretical system for solving the object matching problems in multi-camera tracking. At present the main object matching methods can be classified as geometric constraints methods and feature recognition methods.

*1) Geometric Constraints Methods*

Geometric constraints methods can be divided into methods based on two-dimensional geometric constraints and methods based on three-dimensional geometric constraints.

- Black et al. [5] proposed a method for object matching by using the human body centroid under the landmark homography matrix constraint. But this method needs the objects be all in the same plane.

- Kelly et al. [6] proposed a method to estimate the object position in the three-dimensional model by multi-camera collaboration, and made the object matching according to the position information estimated. But this method needs to know the three-dimensional knowledge about the environment.

- Liang et al. [7] proposed a head detection and trifocal tensor pointer transfer method for object matching by multi-camera collaboration. In this method, people's head position is detected after background subtraction and tracked by Kalman and PDA. Trifocal tensor transfer is used to locate objects in the virtual top view by the corresponding head points in two camera views.

*2) Feature Recognition Methods*

Feature recognition methods do the feature extraction from the object firstly, and based on the features it builds the appearance models of object before different cameras, and then the transfer model among different appearance models is estimated, finally by estimating the similarity of appearance models the object matching results are determined. The object feature includes color feature, point feature, line feature, local feature, texture, invariant feature et al. In recent researches, the general algorithms choose one or some features in establishing the appearance model for object matching.

- Javed et al. [8] and Kumar et al. [9] built the appearance model and Brightness Transfer Function (BTF) by using the color histogram of the object with color feature, as improved the accuracy of object matching. Prosser et al. [10] proposed a color transfer model based on bidirectional cumulative color histogram, as is better than the BTF. Mazzeo et al. [11] made a comparison of the mean BTF and cumulative BTF, and pointed out that the cumulative BTF is more accurate for mapping of rare brightness.

- RGB color feature is easily affected by illumination. Mazzeo et al. [12] verified the matching accuracy of H component histogram is higher than that of RGB histogram, but he also pointed out after BTF conversion the matching accuracy of RGB color histogram is higher than that of H component histogram.

- In the non-rigid object tracking, the methods with local features are widely used because of their good invariance. The related works can be divided into two kinds; one is based on absolute value and the other is based on relative value. SIFT [13] is the typical method based on absolute value, as constructs a histogram with gray and gradient quantization. SIFT is not only scale invariant, but also can get good detection results when changing the rotation angles, image brightness or shooting angles. The shortcomings of SIFT is it needs a large amount of computation time. PCA-SIFT [14] method replaces the histogram in SIFT by principal component analysis, and improves the computation speed. SURF [15, 16] method has equivalent performance with SIFT, but it computes faster than SIFT. Rublee et al. [17] proposed ORB method with local feature point, as has strong robustness on illumination variation or rotation, and its computation speed is 10 times faster than SURF, but the ORB method is not robust to scale change. The method based on relative value, such as BRIEF [18], OSID [19], BRISK [20], constructs descriptor by comparing the characteristic value of pre-trained or random points, and it has small amount of computation. Detection and description of the local feature points, less computation, and stronger local description ability are still research challenges.

- Texture also has a good effect in object matching. Tuzel et al. [21] proposed a method using a covariance matrix about regional color feature and gradient feature to express the region, and the expression has strong power. In the VIPER database it performs better than HOG, LBP and other methods [22].

- Each feature has its advantages and disadvantages, so the combination of global and local features, or the combination of static and dynamic features, or using the gait, face and other biological characteristics together to express the object can get better matching results. Hirzer et al. [23] realized an across camera matching method with HSV, Lab and LBP feature fusion.

*B. Cameras Calibration and Estimation of Their Topological Relations*

Generally speaking, the aim of Camera Calibration is to obtain the parameters of cameras, with these parameters to obtain the topological relations of cameras, and then obtain the object trajectory.

Set $y_i$ represent the observed trajectory vector of object $O$ in the $i$th camera, $\hat{y}$ represent the real trajectory vector of object $O$, so the trajectory matching function of object $O$ in the whole $C$ cameras can be written as following equation (2):

$$\arg\min_{y_i \in O} \left\| \sum_{i=1}^{C} \|y_i\| - \hat{y} \right\|, \quad y_{i+1} = A_i y_i + w_i \quad (2)$$

$A_i$ represents the evolution path of the object based on topological relations of cameras, and $w_i$ is the Gauss noise with zero mean.

Due to the expansion of multi-camera surveillance area, automatically obtaining topological relations of cameras has become an important part of multi-camera tracking research. The related work can be divided into topology estimation based on feature change and topology estimation based on transit time distribution.

*1) Topology Estimation based on Feature Change*

This kind of methods obtains the topology estimation of cameras with available objects, reliable visual features or motion features, but is easily influenced by the appearance changes of camera view or the attitude of people. Therefore it needs to be supervised or unsupervised learning to get the right relations of cameras.

- Javed et al. [8] constructed object color model with Gauss distribution firstly, and then use the Parzen window by supervised learning method to estimate the probability distribution of transfer time with time interval, entering point position and velocity, and finally match the object by combining all the information, so as to realize the object tracking in multiple cameras.

- Gilbert and Bowden [24] estimated the temporal relations of different cameras and the entering relations of the object showing in the cameras by unsupervised cumulative learning method. Due to using unsupervised learning, the method can adapt to environmental changes, but the learning time is longer.

- Motamed and Wallart [25] represented the occurent probability of observed object in the next camera with fuzzy intervals, and this possibility can be obtained by estimating its motion equations.

*2) Topology Estimation based on Transit time Distribution*

This kind of methods obtains the topology estimation of cameras by constructing the transit time distribution through detecting the object entering and leaving time from the perspectives of different cameras.

- Pasula et al. [26] used online EM methods to learn the transit time distribution under given topology constraints.

- Ellis et al. [27] established the spatio-temporal topological relations of cameras automatically by using unsupervised leaning methods to learn the observation data of object in multi-camera surveillance network, Because the algorithm only considers the temporal and spatial information, it is not subject to limits on the camera characters, the observation directions and other factors. Inspired by the Ellis method, Tieu [28] made a combination of uncertain correspondence and Bayesian methods, and by reducing the assumed conditions designed a more general topology estimation algorithm.

- Hengel et al. [29] assumed that all the cameras had underlying connections, and then removed the impossible connections by observation. Experiment results show that the method has a good effect in topology estimation for large networks of cameras, especially when the learning samples are less.

- Chen et al. [30] by using unsupervised learning method respectively designed the import route model and transit time model of objects based on mixed Gauss model, and the learning process is divided into two stages: a batch learning and an online incremental learning, in order to improving its adaptability.

*C. Information Fusion*

Information fusion aims to combine the appearance characteristics of object with the topology characteristics of the camera, and ultimately realizes the object continuous tracking in the monitoring area. In order to keep the object continuous tracking in the wide multi-camera monitoring area, it must consider the object handover problem between different cameras. The key in handover problem is how to find the next handoff camera, so that the number of handover is the minimum in the tracking process, and at the same time, information redundancy and information loss is least.

Set $z(x_i, x_j)$ be the characteristic function of object $O$, as represents the handover process between the $i$th camera and the $j$th camera. If the handover process is successful, $z(x_i, x_j) = 1$ or $z(x_i, x_j) = 0$, so the information fusion function of object $O$ in the whole $C$ cameras can be written as following equation (3):

$$\min_{x_i, x_j \in O} \sum_{i=1}^{C} \sum_{j=1, j \neq i}^{C} z(x_i, x_j), \quad s.t. \begin{cases} \min_{x_i, x_j \in O, i \neq j} \|x_i - x_j\| \\ \min_{x_i, x_j \in O, i \neq j} \|y_i - y_j\| \end{cases} \quad (3)$$

In addition to detect, unite, and estimate the multi-camera data, sometimes information fusion also dose the sensor signal and priori information fusion. So information fusion is a multi-level, stair-step, multi-source information fusion process. Many classic information fusion framework and methods may be used to solve the problem, for example, methods based on Bayesian estimation model [31], methods base on Kalman filtering [32], methods based on particle filtering [33] et al. According to whether there exists overlapping regions among cameras, information fusion methods can be divided into information fusion with overlapping cameras and non-overlapping cameras.

*1) Information Fusion with Overlapping cameras*

- Munoz-Salinas et al. [34] combined Bayesian filtering with Dempster-Shafer evidence theory, and proposed the evidence filtering method for solving the multi-camera multi-object indoor tracking problem.

- Qu et al. [35] proposed a distributed Bayesian algorithm for object tracking, as can solve the information fusion problem with overlapping cameras.

- Du et al. [36] proposed an object tracking algorithm for multi-camera tracking with overlapping areas, as combined sequential belief propagation with particle filtering algorithm.

- Cai et al. [37] established an index named tracking confidence for each object. When tracking confidence of an object is lower than the threshold, the system began to perform a global search in camera network, and activate the camera, whose tracking confidence related to the object is the highest, for object tracking

*2) Information Fusion with Non Overlapping cameras*

- Kettnaker and Zabih [38] established an object tracking model for multi-camera system based on Bayesian theory. Using the methods used in linear programming problems, the model is solved as a solution of maximum posteriori probability, so as to realize the object tracking.

- Chilgunde et al. [39] established a multi-camera object tracking algorithm based on Kalman filtering. Kalman filtering is used not only to track object within the camera view, but also when the object leaves the camera view it still can track the object. It realizes the object matching between different cameras by constructing Gauss model with shape, movement, location and other characteristics of the object.

- Leoputra et al. [40] applied priori information to the design of multi-camera tracking algorithm, and proposed a particle filtering algorithm for object tracking based on the pre-known camera topology.

- SVM is a new efficient algorithm for recognition [41]. Bauml et al. [42] combined SVM with the DCT features to realize object association. Prosser et al. [43] regarded the person recognition problem as a sequencing problem. By using Ensemble Rank SVM method it realizes the sequencing match, and reduces the computation time of the original SVM method. The above methods regard object association as a general classification problem, while Avraham et al. [44] based on the thought of "couple samples" proposed a new algorithm for object association by learning SVM classifiers. In this method two feature vectors of the same object in two cameras are concatenated into positive samples and two feature vectors of different objects in two cameras are concatenated into negative samples. SVM classifiers are learned based on such data source. The proposed method can effectively reflect the latent difference between two cameras.

Multi-camera tracking faces new technical challenges in contrast to single camera tracking, to solve them many classic theories and algorithms have been used in object matching, cameras calibration and topology estimation, and information fusion. But their application result is not always satisfactory. Table I makes a list of the theories used in multi-camera tracking and shows that there still exist some blank fields need to be further studied.

TABLE I. THEORY USED IN MULTI-CAMERA TRACKING

| Theory | Multi-Camera Tracking | | |
|---|---|---|---|
| | Object Matching | Cameras Calibration and topology estimation | Information Fusion |
| Gauss | √ | √ | √ |
| Bayesian | √ | √ | √ |
| Kalman Filtering | | | √ |
| Particle filtering | | | √ |
| SVM | √ | | √ |

III. CHALLENGE OF SYSTEM ARCHITECTURE

System architecture design is another challenge of multi-camera tracking. The traditional system architecture of multi-camera surveillance system is the centralized architecture, video data acquired by cameras are sent directly to the control center, data alignment, object matching, track record, information fusion and post processing (integrated tracking, et al.) are all done in control center, and the camera itself has no independent processing ability (Shown in Fig. 2).

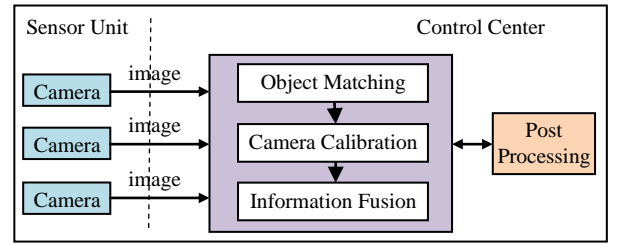

Fig. 2. Centtralized architectrue of multi-camera tracking

The characteristics of this architecture are less information loss, high tracking precision, but high system communication ability, heavy computational load for control center, low efficiency, and poor real-time performance. So in detection and identification of large-scale scene its analysis ability is very limited, it is necessary to build new system architecture to meet the application requirements with large data. The current research on the system architecture are focused on building distributed system architecture, which can be divided into distributed architecture based on camera processor [45,46] and distributed architecture based on object agent [47,48].

*A. Distributed Architecture based on CameraProcessor*

The distributed architecture based on camera processor consists of Sensor Processing Unit (SPU), Central Processing Unit (CPU) and Post Processing (PP). (Shown in Fig. 3)

*1) Sensor Processing Unit*

SPU is composed of some modules of Camera Processor (CP). Each CP not only has the camera function, but also has the independent processing ability, and can automatically obtain the video data, object detection, classification, tracking etc. within itself. Furthermore, it can pass the computing results to the CPU, such as the object type, position, speed, time stamp, camera parameters (displacement, rotation, amplification) and other metadata.

*2) Central Processing Unit*

CPU mainly completes the information fusion between CPs, establishing communication between CPs, and the database operations of related information. Allocation and scheduling of CPs is the key function of CPU, as is done according to task priority, burden of SPU, camera visibility, and other affecting factors.

*3) Post Processing*

PP mainly includes Graphic User interface (GUI). By interaction with users it can realize integrated tracking, constrained tracking, path prediction and other advanced applications

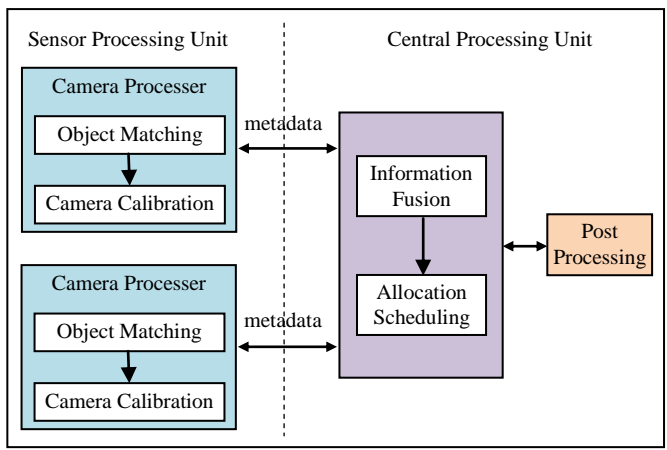

Fig. 3. Distributed Architecture based on Camera Processor

B. *Distributed Architecture based on Object Agent*

The distributed architecture based on object agent adopts the object oriented multi-camera structure, and it mainly consists of Detection Cluster Unit (DCU), Processing Cluster Unit (PCU), Cluster Manager (CM) and Post Processing (PP). (Shown in Fig. 4)

*1) Detection Cluster Unit*

DCU is composed of some Detection Agents (DA). DA does the similar function as the Camera Processor (CP in Fig.3), but it adds the function of interaction with the PCU.

*2) Processing Cluster Unit*

PCU is composed of some Object Agents (OA), and each OA corresponds to an object tracking. According to the object status, the parameters of DA and the image quality estimated, PCU allocates and schedules the OAs to dynamically control multiple DAs for object tracking. PCU is the dynamic layer between DCU and CM, and is the most complex processing unit.

*3) Cluster Manager*

CM is responsible for managing the dynamic layer PCU. CM allocates the OA with optimal parameters for each object, and users can set parameters of PCU via CM.

*4) Post Processing*

PP performs the same function as it does in Fig.3.

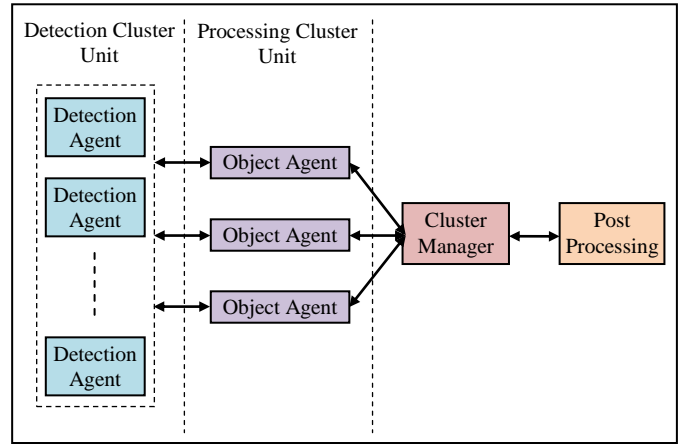

Fig. 4. Distributed Architecture based on Object Agent

IV. CONCLUSION

Multi-camera tracking is an interdisciplinary research problem, and it puts forward new challenges to many traditional techniques and methods. Especially with the improvement of hardware performance and expansion of application area, it ushers in some new problems. From the development trend of technology, the integrated application of multi techniques and multi theories is the key to solve these problems in the future. Meanwhile, from the development trend of the system architecture, distributed system architecture, as preposes the data processing function, improves the computation ability of cameras, reduces the functions of control center and the control center only performs the function of organization and coordination, will be the key to improve the system's processing ability in the future.